\documentclass[sigconf]{acmart}
\usepackage{bm}
\usepackage{amsmath,amsfonts}
\usepackage{algorithm}
\usepackage{fixfoot}
\usepackage[scaled=0.9]{couriers}
\usepackage{algorithmic}
\usepackage{multirow}
\usepackage{latexsym}
\usepackage{booktabs}
\usepackage{graphicx}
\usepackage{enumerate}
\usepackage{subfigure}
\usepackage{mathrsfs}
\usepackage{textcomp}

\AtBeginDocument{%
  \providecommand\BibTeX{{%
    \normalfont B\kern-0.5em{\scshape i\kern-0.25em b}\kern-0.8em\TeX}}}

\setcopyright{iw3c2w3}

\begin{document}
\DeclareFixedFootnote{\evalcode}{\url{https://github.com/allenai/propara/tree/master/propara/evaluation}}

\title{Knowledge-Aware Procedural Text Understanding with Multi-Stage Training}

\author{Zhihan Zhang}
\authornote{Work was done while Zhihan Zhang was an intern at STCA NLP Group, Microsoft.}
\affiliation{%
  \institution{Peking University}
  \city{Beijing}
  \country{China}
}
\email{zhangzhihan@pku.edu.cn}

\author{Xiubo Geng}
\authornote{Corresponding authors.}
\affiliation{%
  \institution{STCA NLP Group, Microsoft}
  \city{Beijing}
  \country{China}
  }
\email{xigeng@microsoft.com}

\author{Tao Qin}
\affiliation{%
  \institution{STCA NLP Group, Microsoft}
  \city{Beijing}
  \country{China}
  }
\email{taoqin@microsoft.com}

\author{Yunfang Wu}
\affiliation{%
  \institution{Peking University}
  \city{Beijing}
  \country{China}
  }
\email{wuyf@pku.edu.cn}

\author{Daxin Jiang}
\authornotemark[2]
\affiliation{%
  \institution{STCA NLP Group, Microsoft}
  \city{Beijing}
  \country{China}
  }
\email{djiang@microsoft.com}

\renewcommand{\shortauthors}{Zhihan Zhang, Xiubo Geng, Tao Qin, Yunfang Wu and Daxin Jiang}

\begin{abstract}
Procedural text describes dynamic state changes during a step-by-step natural process (\textit{e.g.}, photosynthesis). In this work, we focus on the task of procedural text understanding, which aims to comprehend such documents and track entities' states and locations during a process. Although recent approaches have achieved substantial progress, their results are far behind human performance. Two challenges, the difficulty of commonsense reasoning and data insufficiency, still remain unsolved, which require the incorporation of external knowledge bases. Previous works on external knowledge injection usually rely on noisy web mining tools and heuristic rules with limited applicable scenarios. In this paper, we propose a novel \textbf{K}n\textbf{O}wledge-\textbf{A}ware procedura\textbf{L} text underst\textbf{A}nding (\textsc{KoaLa}) model, which effectively leverages multiple forms of external knowledge in this task. Specifically, we retrieve informative knowledge triples from ConceptNet and perform knowledge-aware reasoning while tracking the entities. Besides, we employ a multi-stage training schema which fine-tunes the BERT model over unlabeled data collected from Wikipedia before further fine-tuning it on the final model. Experimental results  on two procedural text datasets, ProPara and Recipes, verify the effectiveness of the proposed methods, in which our model achieves state-of-the-art performance in comparison to various baselines.\footnote{Code is available at \url{https://github.com/ytyz1307zzh/KOALA}}
\end{abstract}


\keywords{Procedural Text Understanding, Entity Tracking, Knowledge-Aware Reasoning, Multi-Stage Training}

\copyrightyear{2021}
\acmYear{2021}
\acmConference[WWW '21]{Proceedings of the Web Conference 2021}{April 19--23, 2021}{Ljubljana, Slovenia}
\acmBooktitle{Proceedings of the Web Conference 2021 (WWW '21), April 19--23, 2021, Ljubljana, Slovenia}
\acmPrice{}
\acmDOI{10.1145/3442381.3450126}
\acmISBN{978-1-4503-8312-7/21/04}
\settopmatter{printacmref=true}

\maketitle

\section{Introduction}
\label{sec:intro}

\begin{figure}[tb]
    \centering
    \includegraphics[width=1.0\linewidth]{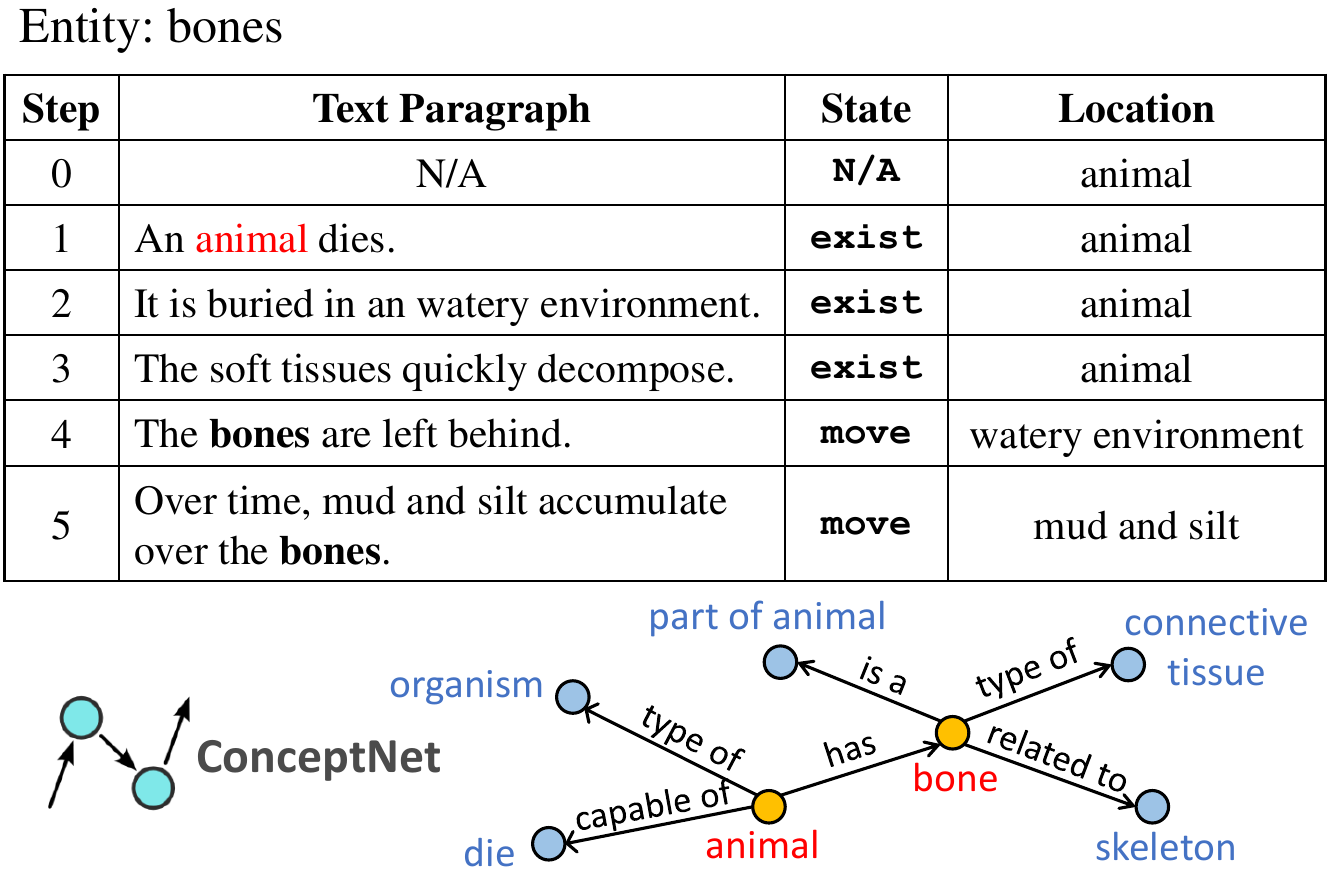}
    \caption{An example of a procedural text paragraph describing fossilization, and the state \& location labels of entity ``bones''. Step 0 is used to identify entities' initial locations before the process. Below is part of the ConceptNet knowledge graph pertaining to the process.}
    \label{fig:task_example}
\end{figure}

In this work, we focus on a challenging branch of natural language processing (NLP), namely procedural text understanding. Procedural text describes dynamic state changes and entity transitions of a step-by-step process (\textit{e.g.}, photosynthesis). Understanding such procedural text requires AI models to track the participating entities throughout a natural process  \cite{dalvi2018propara, bosselut2018npn}. Taking Figure \ref{fig:task_example} for example, given a paragraph describing the process of fossilization and an entity ``bones'', the model is asked to predict the \textit{state} (not exist, exist, move, create or destroy) and \textit{location} (a textspan from the paragraph) of the entity at each timestep. Such procedural texts usually include the comprehension of underlying dynamics of the process, thus impose higher requirements on the reasoning ability of NLP systems.

Since the proposal of the procedural text understanding task \cite{dalvi2018propara}, many models have emerged to solve this challenging task. Recent approaches usually focus on designing effective task-specific reading comprehension models to dynamically encode the changing world of procedural texts and achieve competitive results \cite{tang2020ien, gupta2019ncet, amini2020dynapro}. However, the highest result so far (\textasciitilde 65 F1) are still behind human performance (83.9 F1). Particularly, there are two major problems that has not been effectively solved in this task.

First, commonsense reasoning plays a critical role in understanding procedural text. Without leveraging external knowledge, typical end-to-end models assume that the clues for making predictions have already existed in plain text, which does not always hold in this task. Not only do entities usually undergo implicit state changes, but their locations are also omitted in many cases, especially when humans can easily infer the location through commonsense reasoning. For instance, in the example in Figure \ref{fig:task_example}, due to the decoupling of the entity ``bones'' and location ``animal'' in the paragraph, the initial location of ``bones'' is hard to be directly inferred from plain text, unless the model is aware of extra commonsense knowledge \textit{``bones are parts of an animal''}. For statistical evidence, we manually check 50 instances from the popular ProPara dataset \cite{dalvi2018propara}. Among these samples, we find that an entity is not explicitly connected to its locations in 32\% of the cases, and state changes (create/move/destroy) of an entity are not explicitly stated in 26\% of the cases. These figures suggest that the need of commonsense knowledge is unneglectable for understanding procedural documents.

Second, data insufficiency hinders large neural models from reaching their best performances. Since data annotation on this task includes states and locations of all entities in each timestep, fully annotated data are costly to collect. As a result, existing datasets are limited in size. The benchmark ProPara dataset only contains 488 paragraphs including 1.9k entities. Although another recent dataset, Recipes \cite{bosselut2018npn}, contains 66k paragraphs, only 866 of them have reliable human-annotated labels, while other paragraphs are automatically machine-annotated and contain lots of noise \cite{gupta2019et}. Moreover, such paragraphs usually fail to provide sufficient information considering the complexity of scientific processes. For example, each paragraph in ProPara only contains \textasciitilde60 words on average (see Table \ref{tab: dataset} for more stats), which restricts it from describing a complex process in detail. Thus, data enrichment is in serious need on this task.

Due to the need of additional knowledge in this task, incorporating external knowledge to assist prediction has been an important idea in previous procedural text understanding models. For instance, ProStruct \cite{tandon2018prostruct} writes heuristic rules to constraint the transition of entity states, while using Web text to estimate the probability of an entity undergoing certain state changes. Similarly, XPAD \cite{dalvi2019xpad} also collects Web corpus to estimate the probability of action dependency. However, their approaches have limitations in both forms of knowledge and applicable scenarios. Using unstructured Web text to calculate co-occurrence frequency requires off-the-shelf tools or heuristic rules, which, unfortunately, often induce lots of noise. Besides, such methods are only applicable to refine the probability space of state change prediction, which do not cover location prediction and have poor generalization ability. Different from previous works, in this paper, we aim to effectively leverage both structured and unstructured knowledge for procedural text understanding. Structured knowledge, like relational databases, provides clear and reliable commonsense knowledge compared to web-crawled text. As for unstructured knowledge like Web text, instead of directly mining probability information, we propose to utilize it with a multi-stage training schema on BERT encoders to circumvent potential noise induced by Web search and text mining. Therefore, we propose task-specific methods to effectively leverage multiple forms of knowledge, both structured and unstructured, to help neural models understand procedural text.

Based on such motivation, we aim to address the above two issues, commonsense reasoning and data inefficiency, using external knowledge sources, namely ConceptNet and Wikipedia. To solve the challenge of commonsense reasoning, we perform knowledge infusion using ConceptNet \cite{speer2017conceptnet}. Consisting of numerous (\texttt{subject,relation,object}) triples, ConceptNet is a relational knowledge base composed of concepts and inter-concept relations. Such structure makes ConceptNet naturally suitable for entity-centric tasks like procedural text understanding. An entity in our task can be matched to a concept-centric subgraph in ConceptNet, including its relations with neighboring concepts. Such information can be used as extra commonsense knowledge to help models understand the attributes and properties of an entity, which further provides clues for making predictions even if the answers are not directly mentioned in plain text. As shown in Figure \ref{fig:task_example}, although it is hard to directly infer the initial location of ``bones'', we can find triples (\texttt{animal,HasA,bone}) and (\texttt{bone,IsA,part\char`_of\char`_animal}) from the ConceptNet knowledge graph. These knowledge triples can serve as evidence for predicting entity states and locations that are not explicitly mentioned. Therefore, we propose to retrieve relevant knowledge triples from ConceptNet, and apply attentive knowledge infusion to our model, which is further guided by a task-specific attention loss. 

As for the challenge of data insufficiency, we propose to enrich the training procedure using Wikipedia paragraphs based on text retrieval. Inspired by the great success of ``pre-train then fine-tune'' procedure of BERT models \cite{devlin2019bert}, we propose a multi-stage training schema for BERT encoders. Specifically, we simulate the writing style of procedural text to retrieve similar paragraphs from Wikipedia. Compared to paragraphs in existing datasets, such Wiki paragraphs are usually longer, more scientific procedural texts and contain more details about similar topics. We expect the BERT model learn to better encode procedural text through fine-tuning on this expanded procedural text corpus. Thus, we train the BERT encoder for an additional language modeling fine-tuning phase with modified masked language model (MLM) objective, before further fine-tuning the whole model on the target dataset. We also conduct a similar multi-stage training schema on ConceptNet knowledge modeling where we adopt another BERT encoder.

Based on the above approaches, we introduce our \textbf{K}n\textbf{O}wledge-\textbf{A}ware procedura\textbf{L} text underst\textbf{A}nding (\textsc{KoaLa}) model, which effectively incorporates knowledge from external knowledge bases, ConceptNet and Wikipedia. \textsc{KoaLa} infuses commonsense knowledge from ConceptNet during decoding and is trained with a multi-stage schema using an expanded corpus from Wikipedia. For evaluation, our main experiments on ProPara dataset show that \textsc{KoaLa} reaches state-of-the-art results. Besides, auxiliary experiments on Recipes dataset also demonstrate the advantage of our model over strong baselines. The ablation tests and case studies further show the effectiveness of the proposed methods, which makes \textsc{KoaLa} a more knowledgeable procedural text ``reader''.

The main contributions of this work are summarized as follows.

\begin{itemize}
\item We propose to apply structured knowledge, ConceptNet triples, to satisfy the need of commonsense knowledge in understanding procedural text. Knowledge triples are extracted from the ConceptNet knowledge graph and incorporated into an end-to-end model in an attentive manner. A task-specific attention loss is introduced to guide knowledge selection.
\item We propose to use unstructured knowledge, Wikipedia paragraphs, to address the issue of data inefficiency in this task. Through a multi-stage training procedure, the BERT encoder is first fine-tuned on retrieved Wiki paragraphs using task-specific training objectives before further fine-tuned with the full model on the target dataset.
\item Experimental results show that our knowledge-enhanced model achieves state-of-the-art results on two procedural text datasets, ProPara and Recipes. Further analyses prove that by effectively leveraging external knowledge sources, the proposed methods helps the AI model better understand procedural text.
\end{itemize}

\section{Related Work}
\label{sec:related_work}
\paragraph{Procedural Text Datasets} Efforts have been made towards researches in procedural text understanding since the era of deep learning. Some earlier datasets include bAbI \cite{weston2016babi}, SCoNE \cite{long2016scone} and ProcessBank \cite{berant2014processbank}. bAbI is a relatively simple dataset which simulates actors manipulating objects and interacting with each other, using machine-generated text. SCoNE aims to handle ellipsis and coreference within sequential actions over simulated environments. ProcessBank consists of text describing biological processes and asks questions about event ordering or argument dependencies.

In this paper, we mainly focus on ProPara \cite{dalvi2018propara}, a more recent dataset containing paragraphs on a variety of natural processes. The goal is to track states and locations of the given entities at each timestep. Additionally, we also conduct experiments on Recipes dataset \cite{bosselut2018npn}, which includes entity tracking in the cooking domain. These datasets are more challenging since AI models need to track the dynamic transitions of multiple entities throughout the process, instead of predicting the final state (SCoNE) or answer a single question (bAbI, ProcessBank). Besides, entities usually undergo implicit state changes and commonsense knowledge is often required in reasoning.

\paragraph{Procedural Text Understanding Models}

Our paper is mainly related to the lines of work on ProPara \cite{dalvi2018propara}. ProStruct \cite{tandon2018prostruct} applies VerbNet rulebase and Web search co-appearance to refine the probability space of entity state prediction. LACE \cite{du2019lace} introduces a consistency-biased training objective to improve label consistency among different paragraphs with the same topic. KG-MRC \cite{das2019kgmrc} constructs knowledge graphs to dynamically store each entity's location and to assist location span prediction. NCET \cite{gupta2019ncet} extracts candidate locations using part-of-speech rules from text paragraphs, and considers location prediction as a classification task over the candidate set. ET \cite{gupta2019et} conducts analyses on the application of pre-trained BERT and GPT models on the sub-task of state tracking. XPAD \cite{dalvi2019xpad} builds dependency graphs on ProPara dataset, which tries to explain the action dependencies within the events happened in a process. Among more recent approaches, \textsc{Dynapro} \cite{amini2020dynapro} dynamically encodes procedural text through a BERT-based model to jointly identify entity attributes and transitions. ProGraph \cite{zhong2020prograph} constructs an entity-specific heterogeneous graph on temporal dimension to assist state prediction from context. IEN \cite{tang2020ien} explores inter-entity relationship to discover the causal effects of entity actions on their state changes. In this paper, we aim at two main problems that have not been effectively solved by the above works: commonsense reasoning and data insufficiency. Benefiting from the commonsense knowledge in ConceptNet and the proposed multi-stage training schema, our model outperforms the aformentioned models on the ProPara dataset.

\paragraph{Commonsense in Language Understanding}

Incorporating commonsense knowledge to facilitate language understanding is another related line of work \cite{storks2019survey, yu2020survey}. Yang et al. \cite{yang2017leverage} infuse concepts from WordNet knowledge base with LSTM hidden states to assist information extraction. Chen et al. \cite{chen2018inference} propose a knowledge-enriched co-attention model for natural language inference. Lin et al. \cite{lin2019kagnet} employ graph convolutional networks and path-based attention mechanism on knowledge graphs to answer commonsense-related questions. Guan et al. \cite{guan2019storyend} apply multi-source attention to connect hierarchical LSTMs with knowledge graphs for story ending generation. Min et al. \cite{min2019knowledge} construct relational graph using Wikipedia paragraphs to retrieve knowledge for open-domain QA. Wang et al. \cite{wang2020adapter} inject factual and linguistic knowledge into language models by training multiple adapters independently. Inspired by previous works, we introduce commonsense knowledge from ConceptNet \cite{speer2017conceptnet} into the procedural text understanding task, and prove that the retrieved knowledge contributes to the strong performance of our model.

\section{Problem Definition}

\begin{figure*}[ht]
  \centering
  \includegraphics[width=1.0\textwidth]{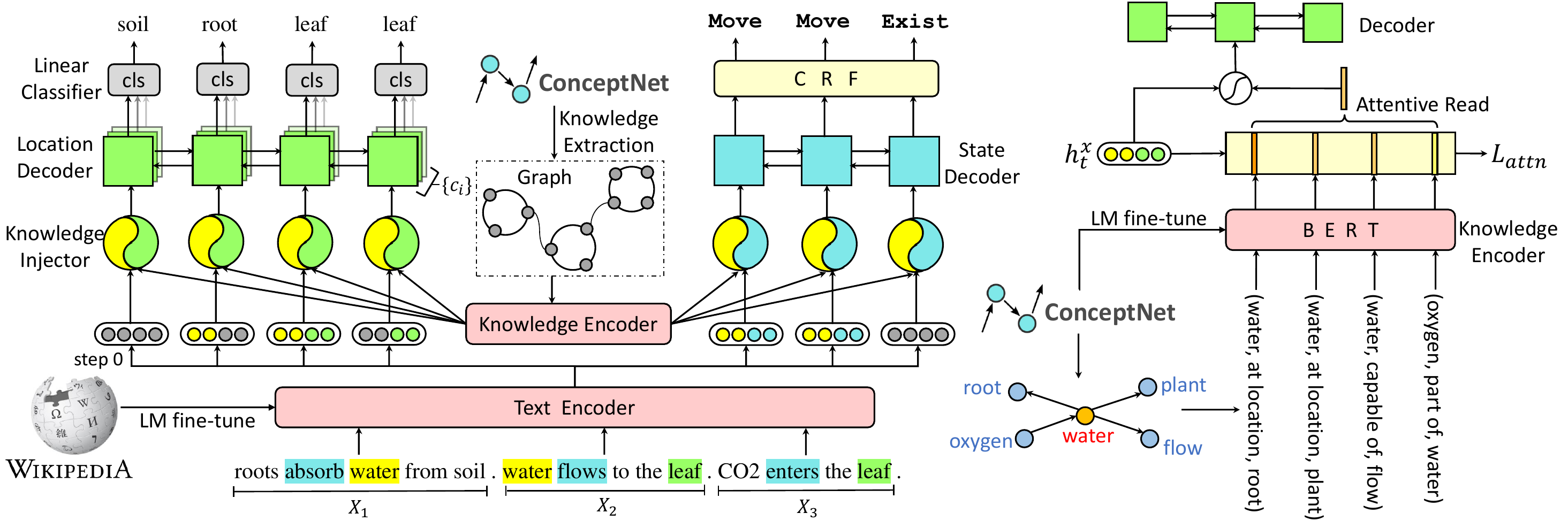}  
\caption{An overview of the \textsc{KoaLa} model (left) \& a detailed illustration of knowledge-aware reasoning modules (right), focusing on entity ``water''. Note that the location prediction modules are applied to each location candidate (root, soil, leaf, \textit{etc}) in parallel, and perform classification among candidates at each timestep. Text \& knowledge encoders are implemented using BERT. ``Decoder'' represents either the state decoder or the location decoder.}
\label{fig:model}
\end{figure*}

Here we define the task of Procedural Text Understanding. \textbf{Given:}

\begin{itemize}

\item   A \textit{paragraph} $P$ composed of $T$ sentences $(X_1, \cdots, X_T)$, representing a process of $T$ timesteps, \textit{e.g.}, photosynthesis or a cooking recipe.

\item  A set of $N$ pre-given \textit{entities} $\{e_1, \cdots, e_N\}$, which are participants of the process.

\end{itemize}

For each entity $e$, \textbf{Predict:}

\begin{itemize}

\item  The entity's \textit{state} at each timestep $y^s_t\ (1\leq t\leq T)$. For ProPara task, $y^s \in$ \{\texttt{\textbf{not\char`_exist}} \rm{(O)}, \texttt{\textbf{exist}} \rm{(E)}, \texttt{\textbf{move}} \rm{(M)}, \texttt{\textbf{create}} \rm{(C)}, \texttt{\textbf{destroy}} \rm{(D)}\}; for Recipes task, $y^s \in$ \{\texttt{\textbf{absence}}, \texttt{\textbf{presence}}\}.

\item The entity's \textit{location} at each timestep $y^l_t\ (0\leq t\leq T)$, which should be a text span in the paragraph. A special `?' token indicates the entity's location is unknown. $y^l_0$ denotes the initial location before the process begins.

\end{itemize}

Besides, the ground-truth location and state at timestep $t$ are denoted as $\widetilde{y}^l_t$ and $\widetilde{y}^s_t$, respectively. In this paper, we will use $\mathbf{W}$ and $\bm{b}$ to represent trainable weight and bias, respectively.

\section{Model}
In this section, we first present the overview of our model. Then, we describe our procedural text understanding model in detail, followed by the proposed knowledge-aware reasoning methods.

\subsection{Overview}

The base framework of \textsc{KoaLa} is built upon the previous state-of-the-art model NCET \cite{gupta2019ncet}, shown in Figure \ref{fig:model}. Its major differences to NCET are the use of powerful BERT encoders, the knowledge-aware reasoning modules (Section \ref{sec:knowledge}) and the multi-stage training procedure (Section \ref{sec:multi-stage}). Based on an encoder-decoder architecture, the model performs two sub-tasks in parallel: \textit{state tracking} and \textit{location prediction}. A text encoder is first used to obtain the contextualized representations of the input paragraph. Then, two decoders are responsible for tracking the state and location changes of the given entity. Commonsense knowledge extracted from ConceptNet is integrated in the decoding process in an attentive manner. The final training objective is to jointly optimize state prediction, location prediction and knowledge selection.

\subsection{Framework}

\paragraph{Text Encoder \& Knowledge Extraction}
Given a paragraph $P$ and an entity $e$, we first concatenate all $T$ sentences in the paragraph into a single text sequence [CLS]$X_1X_2\cdots X_T$[SEP], where [CLS] and [SEP] are special input tokens of BERT. We then encode the text paragraph using a pre-trained BERT model to obtain the contextual embeddings of each text token. Meanwhile, we extract knowledge triples which are relevant to the entity $e$ and paragraph $P$ from ConceptNet. These triples are encoded by another BERT encoder for their representations, which we will elaborate in Section \ref{sec:knowledge}.

\paragraph{State Tracking Modules}
An entity's state changes are usually indicated by verbs. Therefore, for each sentence $X_t$, we concatenate the contextual embeddings of the entity $\bm{h}^e_t$ and the verb $\bm{h}^v_t$  as input $\bm{h}^s_t$ to the state tracking modules. If the entity is a multi-word phrase or there are multiple verbs in the sentence, we average their embeddings. If the entity does not appear in sentence $X_t$, we set $\bm{h}^s_t$ to a all-zero vector:
\begin{equation}
    \bm{h}^s_t = 
    \begin{cases}
      [\bm{h}^e_t;\ \bm{h}^v_t], & \text{if}\ e \in X_t \\
      \quad\bm{0},&  \text{otherwise}
    \end{cases}
\end{equation}
The state tracking modules include a knowledge injector, a Bi-LSTM state decoder and a conditional random field (CRF) layer. The knowledge injector infuses the extracted ConceptNet knowledge with $\bm{h}^s_t$ as the input to the Bi-LSTM decoder. The Bi-LSTM state decoder acts on the sentence level and models the entity's state at each timestep $t$, which simulates the dynamic changes of entity states on the temporal dimension:
\begin{equation}
    \bm{o}^s_t = [\ \overrightarrow{{\rm LSTM}}(\bm{o}^s_{t-1}, \bm{h}^s_t);\ \overleftarrow{{\rm LSTM}}(\bm{o}^s_{t+1}, \bm{h}^s_t)\ ]
\end{equation}
where $\bm{o}^s_t$ denotes the hidden state of the deocder at timestep $t$, and semicolon denotes vector concatenation. Finally, the CRF layer is applied to compute the conditional log likelihood of ground-truth state sequence $\bm{\widetilde{y}}^s$ and the state loss $L_{state}$ is computed as:
\begin{align}
\mathcal{P}(\bm{y}^s| P, e, G)& \propto {\rm exp}\bigg(\ \sum_{t=1}^{T}\ \big(\mathbf{W}_s\bm{o}^s_t + \psi(y^s_{t-1},y^s_t)\big)\bigg)\\
\vspace{1cm}
L_{state} = &-\frac{1}{T}{\rm log}\mathcal{P}(\bm{y}^s=\bm{\widetilde{y}}^s| P, e, G)
\end{align}
where $G$ denotes the knowledge graph extracted from ConceptNet, which will be elaborated in Section \ref{sec:knowledge}; $\psi(y^s_{t-1},y^s_t)$ is the transition potentials between state tags, which is obtained from CRF's transition score matrix.

\paragraph{Location Candidates}
Predicting the entity's location equals to predicting a text span from the input paragraph. Inspired by \cite{gupta2019ncet}, we split this objective into two steps. We first extract all possible location spans as location candidates $\{c_1, \cdots, c_M\}$ from the paragraph, then perform classification on this candidate set. Specifically, we use an off-the-shelf POS tagger \cite{akbik2018flair} to extract all \textit{nouns} and \textit{noun phrases} as location candidates. Such heuristics reach a 87\% recall of the ground-truth locations on the ProPara test set. We additionally define a learnable vector for location `?', which acts as a special candidate location. 

\paragraph{Location Prediction Modules}
Similar to state tracking, for each location candidate $c_j$ at each timestep $t$, we concatenate the contextual embeddings of the entity $\bm{h}^e_t$ and the location candidate $\bm{h}^c_{j,t}$ as the input $\bm{h}^l_{j,t}$ to the location prediction modules. If the entity $e$ or the location candidate $c_j$ does not appear in sentence $X_t$, we replace it with an all-zero vector instead:
\begin{equation}
    \bm{h}^l_{j,t} = 
    \begin{cases}
      [\bm{h}^e_t;\ \bm{h}^c_{j,t}], & \text{if}\ e \in X_t \ \text{and}\ c_j \in X_t \\
      [\bm{h}^e_t;\ \bm{0}], & \text{if}\ e \in X_t\ \text{and}\ c_j \notin X_t \\
      [\bm{0};\ \bm{h}^c_{j,t}], & \text{if}\ e \notin X_t\ \text{and}\ c_j \in X_t \\
      \quad\bm{0},&  \text{otherwise}
    \end{cases}
\end{equation}
Similar to the state tracking modules, the location prediction modules include a knowledge injector and a Bi-LSTM location decoder followed by a linear classifier. The sentence-level Bi-LSTM location decoder models the entity's location at each timestep $t$, which simulates the dynamic changes of entity locations on the temporal dimension. Since there are $M$ location candidates in total, the location decoder is executed for $M$ times. For each candidate $c_j$ at each timestep $t$, the linear layer outputs a score $\bm{o}^l_{j,t}$ based on the decoder's hidden states:
\begin{equation}
    \bm{o}^l_{j,t} = [\ \overrightarrow{{\rm LSTM}}(\bm{o}^l_{j,t-1}, \bm{h}^l_{j,t});\ \overleftarrow{{\rm LSTM}}(\bm{o}^l_{j,t+1}, \bm{h}^l_{j,t})\ ]
\end{equation}
The scores of all location candidates at the same timestep are normalized using Softmax. Then the location loss $L_{loc}$ is computed as the negative log likelihood of the ground-truth locations:

\begin{align}
   \mathcal{P}(y^l_t&\ |\ P, e, G)={\rm softmax}(\mathbf{W}_l\{\bm{o}^l_{j,t}\}^{M}_{j=1}) \\
    L_{loc} &= -\frac{1}{T}\sum_{t=1}^{T} {\rm log}\mathcal{P}(y^l_t=\widetilde{y}^l_t|P, e,G)
\end{align}

\begin{figure}[b]
    \centering
    \includegraphics[width=1.0\linewidth]{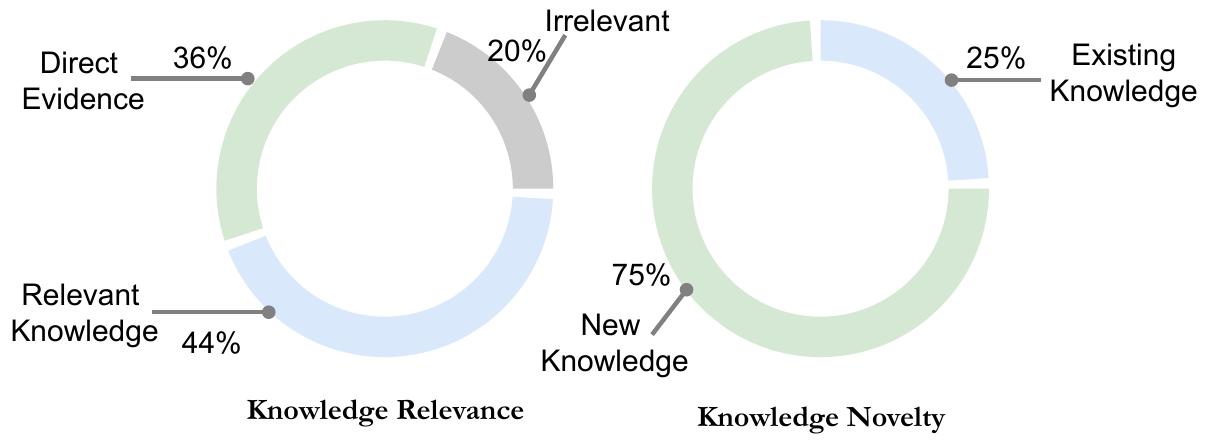}
    \caption{Left: the relevance of the retrieved ConceptNet knowledge to the input paragraph. Right: the novelty of the retrieved knowledge when ConceptNet triples provide useful knowledge.}
    \label{fig:know_extract}
\end{figure}

At inference time, we perform both sub-tasks, but only predict the entity's location when the model predicts its state as \texttt{\textbf{create}} or \texttt{\textbf{move}}, because other states will not alter the entity's location.

\subsection{Knowledge-Aware Reasoning}
\label{sec:knowledge}

Next, we explain the details of injecting ConceptNet knowledge into \textsc{KoaLa}. We first extract those knowledge triples that are relevant to the given entity and input paragraph. Then, we encode these knowledge triples using a BERT encoder. The model attentively reads the knowledge triples and select the most relevant ones to the current context. Additionally, we add a task-specific attention loss to guide the training of knowledge selection modules.

\subsubsection{ConceptNet Knowledge Extraction}

As a large relational knowledge base, ConceptNet is composed of numerous concepts and inter-concept relations. Each knowledge piece in ConceptNet can be regarded as a (\texttt{h,r,t;w}) triple, which means head concept \texttt{h} has relation \texttt{r} with tail concept \texttt{t} and \texttt{w} is its weight in the ConceptNet graph. For a given entity $e$, we first retrieve the entity-centric one-hop subgraph from ConceptNet, \textit{i.e.}, entity $e$ and its neighboring concepts. For phrasal entities that contain multiple words, we retrieve those subgraphs where the central concept $c$ and the entity $e$ has Jaccard Similarity $J(c, e) \geq 0.5$. These subgraphs include the commonsense knowledge related to entity $e$.

Then, we adopt two methods to retrieve relevant triples from this subgraph:

\begin{itemize}

\item Exact-match: the neighboring concept appears in the paragraph $P \rightarrow \{K_e\}$

\item Fuzzy-match: the neighboring concept is semantically related to a content word in the paragraph $P$, according to contextual word embeddings $\rightarrow \{K_f\}$. 

\end{itemize}
where $\{K_e\}$ and $\{K_f\}$ are sets of triples, sorted by weight \texttt{w} and semantic relevance\footnote{The highest embedding similarity between the neighboring concept and any content word in $P$.}, respectively. We select the top $N_K$ triples so that $|\{K_e\}| + |\{K_f\}| = N_K$, while priortizing exact-match ones. The detailed retrieval algorithm is presented in Algorithm \ref{alg: retrieval}. We set $N_K = 10$ in practice.

To testify the efficacy of knowledge extraction, we manually evaluate 50 instances from the ProPara dataset. The results are shown in Figure \ref{fig:know_extract}. Regarding the relevance of the retrieved knowledge, in 36\% of the cases, the knowledge triples provide direct evidence for predicting the entity's state/location; in another 44 \% of the cases, the knowledge triples contain relevant knowledge that helps understand the entity and the context; while the retrieved triples have no relationship with the context in only 20\% of the cases. Among the first two categories, 75\% of the instances can obtain new knowledge that is not indicated in the text paragraph, which verifies the novelty of the retrieved knowledge. These results suggest that the retrieved ConceptNet knowledge is very likely to be helpful from human perspectives.

\subsubsection{Attentive Knowledge Infusion}
\label{sec: attention}

The external knowledge is injected into our model in an attentive manner before the decoders\footnote{Here, ``decoder'' refers to either the state decoder or the location decoder.}, as shown in the right part of Figure \ref{fig:model}. We first encode the ConceptNet triples using BERT. The BERT inputs are formatted as [CLS]\texttt{head}[SEP]\texttt{relation}[SEP]\texttt{tail}[SEP], where \texttt{relation} is interpreted as a natural language phrase. Such formatting scheme converts the original triple into a text sequence while reserving its structural feature. In Section \ref{sec:train_cpnet}, we will describe the multi-stage training procedure which trains BERT encoder to better model ConceptNet triples. We use the average of BERT outputs (excluding [CLS] and [SEP] tokens) as the representation of a knowledge triple:

\begin{algorithm}[t]
\caption{Knowledge retrieval on ConceptNet}
\label{alg: retrieval}
\begin{algorithmic}[1]
\REQUIRE Entity-centric subgraph $G$ composed of $N_G$ triples $\{\tau_1, \cdots, \tau_{N_G}\}$, Paragraph $P$ composed of $N_P$ non-stopword tokens $\{w_1,\cdots,w_{N_P}\}$, entity $e$
\
\STATE $K_e\leftarrow\varnothing, K_f\leftarrow\varnothing$
\FOR{$\tau_i=(e, r_i, n_i; w_i)$ in $G$}
\STATE \texttt{//exact match}
\IF{(WordLen($n_i$)==1 \textbf{and} $n_i$ in $P$) \textbf{or} (WordLen($n_i$)$>$1 \textbf{and} $\frac{\{n_i\}\cap P}{\{n_i\}}\geq 0.5$)}
\STATE $K_e \leftarrow K_e\cup \{\tau_i\}$; \textbf{continue}
\ENDIF
\STATE \texttt{//fuzzy match}
\STATE Generate pseudo-sentence $p^{\tau}_i$ from $\tau_i$\footnotemark
\STATE $\bm{h}^{\tau}_i$ = BERT($p^{\tau}_i$), $\bm{h}^P$ = BERT($P$) 
\STATE $s^{\tau}_i$ = max([cos($\bm{h}^n_i$, $\bm{h}^w$) \textbf{for} $w$ in $P$])
\STATE $K_f \leftarrow K_f\cup \{\tau_i\}$
\ENDFOR
\STATE \texttt{//sort and select} $\ N_K\ $ \texttt{triples}
\STATE sort $K_e$ by $w_i$, sort $K_f$ by $s^{\tau}_i$
\IF{$|K_e|\geq N_K$}
\RETURN top $N_K$ triples in $K_e$
\ELSE
\RETURN $K_e \cup\{$ top $(N_K-|K_e|)$ triples in $K_f\}$
\ENDIF
\end{algorithmic}
\end{algorithm}
\footnotetext{For instance, (\texttt{leaf,PartOf,plant}) can be transformed to ``leaf is a part of plant.''}

\begin{equation}
    \bm{h}^{\tau}_i = {\rm MeanPooling}({\rm BERT}([\texttt{h}, \texttt{r}, \texttt{t}]))
\end{equation}

In order to select the most relevant knowledge to the text paragraph, we use the decoder input as query to attend on the retrieved ConceptNet triples:

\vspace{-0.5cm}
\begin{align}
&\bm{g}^x_t = \sum_{i=1}^{N_K}{\alpha_{i,t}\bm{h}^{\tau}_i} \\
&\alpha_{i,t} =  \frac{exp(\beta_{i,t})}{\sum\limits_{k=1}^{N_K}exp(\beta_{k,t})} \\
&\beta_{i,t} =  \bm{h}^x_t\mathbf{W}_{\beta}{(\bm{h}^{\tau}_i)}^T
\end{align}
where $x\in \{s, l\}$ and $\bm{g}^x_t$ is the graph representation of the retrieved one-hop ConceptNet graph. Finally, we equip the decoder with an input gate to select information from the original input and the injected knowledge:

\begin{align}
\bm{i}^x_t = & \sigma(\mathbf{W}_i[\bm{h}^x_t;\bm{g}^x_t]+\bm{b}_i)\\
\bm{f}^x_t = & \mathbf{W}_f[\bm{h}^x_t;\bm{g}^x_t]+\bm{b}_f\\
{\bm{h}^x_t}' = & \bm{i}^x_t\odot\bm{f}^x_t + (\bm{1}-\bm{i}^x_t)\odot\bm{h}^x_t
\end{align}
where $\odot$ indicates element-wise multiplication and $\sigma$ denotes the sigmoid function. We empirically find that such gated integration performs better than simply concatenating $\bm{h}^x_t$ and $\bm{g}^x_t$ together.

\subsubsection{Attention Loss on Knowledge Infusion}
\label{sec:attn_loss}

Although the attention mechanism can help the model attend on knowledge relevant to the \textit{context}, it is still challenging in some cases to find the most useful triple to the \textit{prediction target} (\textit{i.e.}, the target state and location of the entity). In order to assist the model in learning the dependency between the prediction target and knowledge triples, we use an attention loss as explicit guidance. We heuristically label a subset of knowledge triples that are relevant to the prediction target, and guide the model to attend more on these labeled triples. Recall that we use $\widetilde{y}^l_t$ and $\widetilde{y}^s_t$ to denote the ground-truth location and state of the entity at timestep $t$.

A knowledge triple $\tau_i$ is labeled as 1 (``relevant'') at timestep $t$ if:

\begin{itemize}
\item $\widetilde{y}^l_t \in \tau_i$ and $\widetilde{y}^s_t \in \{\texttt{\textbf{move}}, \texttt{\textbf{create}}\}$, which means the ground-truth location of the current movement/creation is mentioned in $\tau_i$. This is consistent with the inference process in which we only predict a new location when the expected state is \texttt{\textbf{move}} or \texttt{\textbf{create}}.
\item $\tau_i\ \cap\ \mathcal{V}_x \neq \varnothing$ and $\widetilde{y}^s_t = x$, where $x \in \{\texttt{\textbf{move}}, \texttt{\textbf{create}}, \texttt{\textbf{destroy}}\}$. $\mathcal{V}_x$ is the set of verbs that frequently co-appear with state $x$, which is collected from the training set. This suggests that $\tau_i$ includes a verb that usually indicates the occurrence of state change $x$. In practice, we collect those verbs that co-appear with state $x$ for more than 5 times in the training set.
\end{itemize}

Statistically, on the ProPara dataset, 61\% of the data instances have at least one knowledge triples labeled as ``relevant''. On triple-level, 18\% of the knowledge triples are labeled as ``relevant'' for at least once. These figures verifies the trainability of the attention loss since its effect covers a considerable number of training data.

The training objective is to minimize the attention loss, which is to maximize the attention weights of all ``relevant'' triples:

\begin{equation}
L_{attn} = -\frac{1}{N_K\times T}\sum_{i=1}^{N_K}\sum_{t=1}^{T}y^r_{i,t} \cdot{\rm log} \alpha_{i,t}
\end{equation}
where $y^r_{i,t} \in \{0, 1\}$ is the relevance label of triple $\tau_i$ at timestep $t$. Now the model is expected to better identify the relevance between ConceptNet knowledge and prediction target during inference. 

Finally, the overall loss function is computed as the weighted sum of three sub-tasks:

\begin{equation}
\mathcal{L} = L_{state} + \lambda_{loc}L_{loc} + \lambda_{attn}L_{attn}
\label{eq: loss}
\end{equation}
where hyper-parameters $\lambda_{loc}$ and $\lambda_{attn}$ indicate the weights of corresponding sub-tasks in model optimization.

\section{Multi-Stage Training}
\label{sec:multi-stage}

\subsection{Multi-Stage Training on Wikipedia}
\label{sec: train_wiki}

As is mentioned in Section \ref{sec:intro}, we seek to collect additional procedural text documents from Wikipedia to remedy \textit{data insufficiency}. Due to the high cost of human annotation and the unreliability of machine-annotated labels, we adopt self-supervised methods to apply Wiki paragraphs into the training procedure of the text encoder. Inspired by the strong performance of pre-trained BERT models on either open-domain \cite{devlin2019bert} or in-domain data \cite{talmor2019multiqa, xu2019posttrain}, we adopt a multi-stage training schema for the text encoder in our model. Specifically, given the original pre-trained BERT model, we utilize the following training procedure:

\begin{enumerate}[1.]
    \item We perform self-supervised language model fine-tuning (LM fine-tuning) on a procedural text corpus collected from Wikipedia. The training is based on a modified masked language modeling (MLM) objective.
    \item The full \textsc{KoaLa} model, including the BERT encoder, is further fine-tuned on the target ProPara or Recipes dataset.
\end{enumerate}

To collect additional procedural text, for each paragraph $P$ in our target dataset, we split Wiki documents into paragraphs and use DrQA's TF-IDF ranker \cite{chen2017drqa} to retrieve top 50 Wiki paragraphs that are most similar to $P$. Intuitively, we expand the training corpus by simulating the writing style of procedural text. By fine-tuning on a larger corpus of procedural text, we expect the BERT encoder learn to better encode procedural paragraphs on the smaller target dataset. Then, we fine-tune the vanilla BERT on these Wiki paragraphs.

In \textsc{KoaLa}, contextual representations of entities, verbs and location candidates are used for downstream predictions. These tokens are mainly verbs and nouns. Therefore, in order to better adapt the fine-tuned BERT model to the target task, we only apply LM fine-tuning on nouns and verbs. In detail, we observe that nouns and verbs constitute \textasciitilde50\% of all tokens in the collected corpus. To maintain a consistent proportion of masked tokens with BERT's original pre-training \cite{devlin2019bert}, each noun and verb receives a 0.3 mask probability in our MLM objective, whereas the other tokens are never masked. Thus, the fine-tuned BERT is able to generate better representations for nouns and verbs within procedural text corpora.

\begin{figure}[t]
    \centering
    \includegraphics[width=1.0\linewidth]{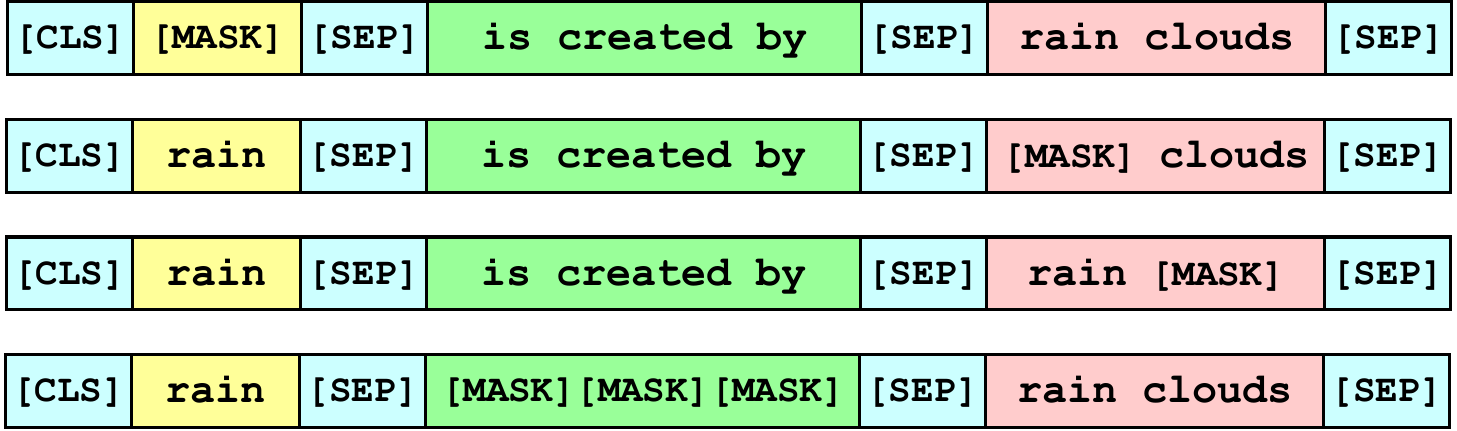}
    \caption{Four instances created from triple (\texttt{rain, CreatedBy, rain\char`_clouds}) in LM fine-tuning on ConceptNet.}
    \label{fig:finetune_cpnet}
\end{figure}

\subsection{Multi-Stage Training on ConceptNet}
\label{sec:train_cpnet}

Inspired by the above fine-tuning schema, we also adopt multi-stage training on the knowledge encoder, which is another BERT model that encodes ConceptNet triples. Different from the text encoder which encodes a sequence of unstructured text, the knowledge encoder models structured ConceptNet triples. Therefore, we modify the conventional MLM objective to fit the structural feature of ConceptNet triples.

Considering the bi-directional architecture of BERT, given a triple $\tau=\ $(\texttt{h,r,t}), we iteratively mask out \texttt{h}, \texttt{r} and \texttt{t} (one at a time) and ask the encoder to predict the masked tokens using the other two unmasked components. Such design assists BERT to better understand the relationships between subjects, relations and objects. However, we empirically find that such masking approach may lead to high information loss and low performance if \texttt{h} or \texttt{t} is too long. Therefore, if \texttt{h} or \texttt{t} consists of more than one tokens, we mask 50\% of the tokens at a time to ensure trainability. We mask all tokens in \texttt{r} since the relation types in ConceptNet is limited (see Figure \ref{fig:finetune_cpnet} for example). The encoder then learns to model the structural information of the knowledge triples through such LM fine-tuning. Similar to Section \ref{sec: train_wiki}, the knowledge encoder is further fine-tuned while \textsc{KoaLa} is trained on the target dataset.

\section{Experiments}
\begin{table}[t]

\begin{tabular}{@{}c|c|cccc@{}}
\toprule
\textbf{Dataset}                  & \textbf{Statistics}                   & \textbf{Train}           & \textbf{Dev}             & \textbf{Test}            & \textbf{Total}           \\ \midrule
\multirow{4}{*}{ProPara} & \#Paragraph                     & 391                      & 43                       & 54                       & 488                      \\
                         & \#Instance                      & 1,504                     & 175                      & 236                      & 1,915                     \\
                         & Avg.sent/para                        & 6.7                      & 6.7                      & 6.9                      & 6.8                      \\
                         & Avg.word/para                        & 61.1                     & 57.8                     & 67.0                     & 61.4                     \\
                         \midrule
\multirow{4}{*}{Recipes} & \#Paragraph & 693  & 86   & 87   & 866  \\
                         & \#Instance  & 5,932 & 756  & 737  & 7,425 \\
                         & Avg.sent/para    & 8.8  & 8.9  & 9.0  & 8.8  \\
                         & Avg.word/para    & 93.1 & 89.1 & 93.9 & 92.8 \\ \bottomrule
\end{tabular}

\caption{Statistics of ProPara and Recipes dataset. The number of instances is equivalent to the total number of entities in all paragraphs.}
\label{tab: dataset}
\end{table}


\subsection{Dataset}

Our main experiments are conducted on the ProPara \cite{dalvi2018propara} dataset\footnote{\url{https://allenai.org/data/propara}}. ProPara is composed of 1.9k instances (one entity per instance) out of 488 human-written paragraphs about scientific processes, which are densely annotated by crowd workers. As an auxiliary task, we also perform experiments on the Recipes \cite{bosselut2018npn} dataset\footnote{\url{http://homes.cs.washington.edu/~antoineb/datasets/nyc_preprocessed.tar.gz}}, which includes cooking recipes and their ingredients. In the original work, human annotation is only applied on the development and test set. Similar to \cite{gupta2019et}, we find that the noise in machine-annotated training data largely lowers models' performances. Therefore, we only use human-labeled data in our experiments and re-split it into 80\%/10\%/10\% for train/dev/test sets. More statistics about these two datasets are shown in Table \ref{tab: dataset}.

\begin{table*}[ht]
\resizebox{0.8\textwidth}!{
\begin{tabular}{l|ccccc|ccc}
\toprule
& \multicolumn{5}{c|}{Sentence-Level} & \multicolumn{3}{c}{Document-Level} \\
\midrule
Models      & Cat-1 & Cat-2 & Cat-3 & Macro-Avg & Micro-Avg & Precision    & Recall    & F1      \\
\midrule
EntNet \cite{henaff2017entnet}     & 51.6  & 18.8  & 7.8   & 26.1      & 26.0      & 54.7         & 30.7      & 39.4    \\
QRN  \cite{seo2017qrn}       & 52.4  & 15.5 &    10.9    & 26.3  & 26.5          &     60.9         &       31.1    &    41.1     \\
ProLocal \cite{dalvi2018propara}   &  62.7           & 30.5           & 10.4           & 34.5               & 34.0    &       \textbf{\underline{81.7}}       & 36.8          & 50.7        \\
ProGlobal \cite{dalvi2018propara}  &     63.0           & 36.4           & 35.9           & 45.1               & 45.4  &      61.7        &    48.8       &    51.9     \\
AQA   \cite{ribeiro2019aqa}      &  61.6  & 40.1  & 18.6  & 39.4  & 40.1      &     62.0         & 45.1          & 52.3        \\
ProStruct \cite{tandon2018prostruct}  &   -    &   -    &  -     &     -      &    -       &      74.3        &   43.0        &   54.5      \\
XPAD  \cite{dalvi2019xpad}      &   -    &    -   &    -   &  -        &    -       &     70.5         &  45.3         &    55.2     \\
LACE   \cite{du2019lace}     &    -   &  -     &  -     &   -        &      -     &     75.3         &   45.4        &   56.6      \\
KG-MRC  \cite{das2019kgmrc}    &    62.9           & 40.0           & 38.2           & 47.0               & 46.6   &     69.3         &    49.3       &  57.6       \\
ProGraph    \cite{zhong2020prograph}    &   67.8   &    44.6 &    41.8  &    51.4  &    51.5   &  67.3 &   55.8     &   61.0   \\
IEN     \cite{tang2020ien}  &   71.8    &    47.6   &   40.5   &    53.3   &    53.0   &   69.8    &   56.3   &   62.3 \\
NCET   \cite{gupta2019ncet}     &  73.7           & 47.1           & 41.0           & 53.9               & 54.0  &         67.1     &     58.5      &   62.5      \\
ET$_{BERT}$ \cite{gupta2019et} &    73.6           & 52.6           & -              & -                  & -                &    -          &      -     &   -      \\
\textsc{Dynapro} \cite{amini2020dynapro}    &  72.4  &   49.3    &   \textbf{\underline{44.5}} &   55.4   & 55.5   &    75.2          &     58.0      &    65.5     \\
\textsc{KoaLa}(Ours) &   \textbf{\underline{78.5}}  & \textbf{\underline{53.3}}  & 41.3  & \textbf{\underline{57.7}}      & \textbf{\underline{57.5}}     &     77.7         &  \textbf{\underline{64.4}}         &     \textbf{\underline{70.4}}   \\
\bottomrule
\end{tabular}
}
\caption{Experiment results on ProPara document-level task and sentence-level task. Most results of the document-level task are collected from the public leaderboard\protect\footnotemark, except for ProGraph and IEN whose scores are self-reported. Results of the sentence-level task are reported by previous works themselves. Some previous approaches did not perform both tasks.}
\label{tab:propara}
\end{table*}
\footnotetext{\url{https://leaderboard.allenai.org/propara/submissions/public}}

\subsection{Implementation Details}

For BERT encoders, we use the BERT$_{BASE}$ model (12-layer transformer with hidden size 768) implemented by HuggingFace's transformers library \cite{wolf2020huggingface}. The whole model contains 235M parameters including 2 BERT encoders. Hyper-parameters in our model are manually tuned according to the model's accuracy on the development set. The parameters during the LM fine-tuning phase in multi-stage training are manually tuned based on the perplexity of BERT encoders. In LM fine-tuning, we set batch size to 16 and learning rate to $5\times 10^{-5}$. The text encoder is trained for 5 epochs on Wikipedia paragraphs, and the knowledge encoder is trained for 1 epoch on ConceptNet triples. While fine-tuning the whole model on target dataset, we use batch size 32 and learning rate $3\times 10^{-5}$ on Adam optimizer \cite{kingba2015adam}. We set $\lambda_{loc}$ to 0.3 and $\lambda_{attn}$ to 0.5 in Eq.(\ref{eq: loss}). Hidden size of LSTMs is set to 256 and the dropout rate is set to 0.4. We train our model for 20 epochs (\textasciitilde1 hour on a Tesla P40 GPU) and select the best checkpoint in prediction accuracy on the development set.

\subsection{Evaluation Metrics}

We perform two tasks, document-level and sentence-level respectively, in our main experiments on ProPara dataset. We perform one task (location change prediction) on Recipes dataset.

\paragraph{Doc-level task on ProPara\evalcode}

Document-level tasks, proposed by \cite{tandon2018prostruct}, require AI models to answer the following document-level questions:

\begin{enumerate}[1.]
    \item What are the \textit{inputs}? The \textit{inputs} are entities that exist at the beginning but are destroyed later in the process.
    \item What are the \textit{outputs}? The \textit{outputs} are entities that are created during the process and exist at the end.
    \item What are the \textit{moves}? The \textit{moves} are times when entities change their locations. The model should predict the old \& new locations of the entity, plus the timestep when the movement occurs.
    \item What are the \textit{conversions}? The \textit{conversions} are times when some entities are destroyed and other entities are created. The model should predict the destroyed \& created entities, plus the location \& the timestep that the conversion occurs.
\end{enumerate}
Evaluation metrics are average precision, recall and F1 scores on the above four perspectives.

\paragraph{Sent-level task on ProPara\evalcode}

Sentence-level tasks, proposed by \cite{dalvi2018propara}, require AI models to answer 3 sets of sentence-level questions:

\begin{enumerate}[1.]
    \item (Cat-1) Is entity $e$ Created (Moved, Destroyed) in the process?
    \item (Cat-2) When (which timestep) is entity $e$ Created (Moved, Destroyed)? 
    \item (Cat-3) Where is entity $e$ Created, (Moved from/to, Destroyed)?
\end{enumerate}
We calculate accuracy for each of the three categories. Evaluation metrics are macro-average and micro-average accuracy of three sets of questions.

\paragraph{Location change prediction on Recipes}

We evaluate our model on the Recipes dataset by how often the model correctly predicts the ingredients' movements, \textit{i.e.}, location changes. For each movement, the model should predict the new location of the entity, plus the timestep when the movement occurs. We report precision, recall and F1 scores on this task.

\subsection{Experiment Results}

In our main experiments on ProPara (Table \ref{tab:propara}), we compare our model with previous works mentioned in Section \ref{sec:related_work}. In the document-level task, \textsc{KoaLa} achieves new state-of-the-art result on F1. Restricted by low recall scores, early approaches struggled in reaching high F1 scores. This could indicate that these models tend to predict less state changes (create/move/destroy), which would result in higher precision and lower recall. In contrary, recent approaches like NCET \cite{gupta2019ncet} and \textsc{Dynapro} \cite{amini2020dynapro} made considerable progress by improving recall performances. Making a step further, our \textsc{KoaLa} model improves document-level comprehension by a large margin. Specifically, \textsc{KoaLa} outscores our base model NCET by 15.8\%/10.1\%/12.6\% on precision/recall/F1 scores, respectively. Compared to the current state-of-the-art model \textsc{Dynapro}, \textsc{KoaLa} achieves 3.3\%/11.0\%/7.5\% relative improvement on precision/recall/F1.

In sentence-level tasks, \textsc{KoaLa} outperforms previous models in most metrics, including state-of-the-art results in macro-average and micro-average scores. The main improvements of \textsc{KoaLa} come from Cat-1 and Cat-2, which means it predicts state changes (create/move/destroy) more accurately than previous models. These results show that \textsc{KoaLa} has stronger ability in modeling procedural text and making predictions on entity tracking.

\begin{table}[t]
\resizebox{1.0\columnwidth}!{
\begin{tabular}{@{}l|ccc@{}}
\toprule
\textbf{Models}              & \textbf{Precision} & \textbf{Recall} & \textbf{F1}   \\ \midrule
NCET \textit{re-implementation}        & 56.5               & 46.4            & 50.9          \\
\midrule
\textsc{KoaLa}                       & \textbf{\underline{60.1}}      & \textbf{\underline{52.6}}   & \textbf{\underline{56.1}} \\
\quad- ConceptNet                & 55.9               & 50.7            & 53.2          \\ 
\quad- LM fine-tuning            & 57.8               & 51.5            & 54.5          \\
\quad- All fine-tuning               & 57.0               & 50.2            & 53.4          \\
\quad- ConceptNet \& fine-tuning & 57.8               & 47.5            & 52.1          \\ \bottomrule
\end{tabular}
}
\caption{Experiment results on re-split Recipes dataset.}
\label{tab: recipes}
\end{table}

\begin{table}[t]
\resizebox{1.0\columnwidth}!{
\begin{tabular}{@{}l|ccc@{}}
\toprule
\textbf{Models}                                    & {\textbf{Precision}} & {\textbf{Recall}}  & {\textbf{F1}}      \\ \midrule
\textsc{KoaLa}                       & { \textbf{\underline{77.7}}}   & { \textbf{\underline{64.4}}} & { \textbf{\underline{70.4}}} \\ \midrule
{\quad - Attention loss}            & { 75.4}            & { 63.8}          & { 69.2}          \\
{\quad - Attention}                 & { 74.2}            & { 63.7}          & { 68.5}          \\
{\quad - ConceptNet}                & { 76.5}            & { 60.7}          & { 67.7}          \\ \midrule
{\quad - LM fine-tuning}            & { 76.7}            & { 62.2}          & { 68.7}          \\
{\quad - All fine-tuning}           & { 73.8}            & { 60.6}          & { 66.5}          \\
{\quad - ConceptNet \& fine-tuning} & { 73.2}            & { 59.2}          & { 65.5}          \\ \bottomrule
\end{tabular}
}
\caption{Ablation tests on ProPara dataset. `` - Attention'' means using average representation of $N_K$ ConceptNet triples instead of using attention to select information.}
\label{tab: ablation}
\end{table}

\begin{figure}[tb]
    \centering
    \includegraphics[width=1.0\linewidth]{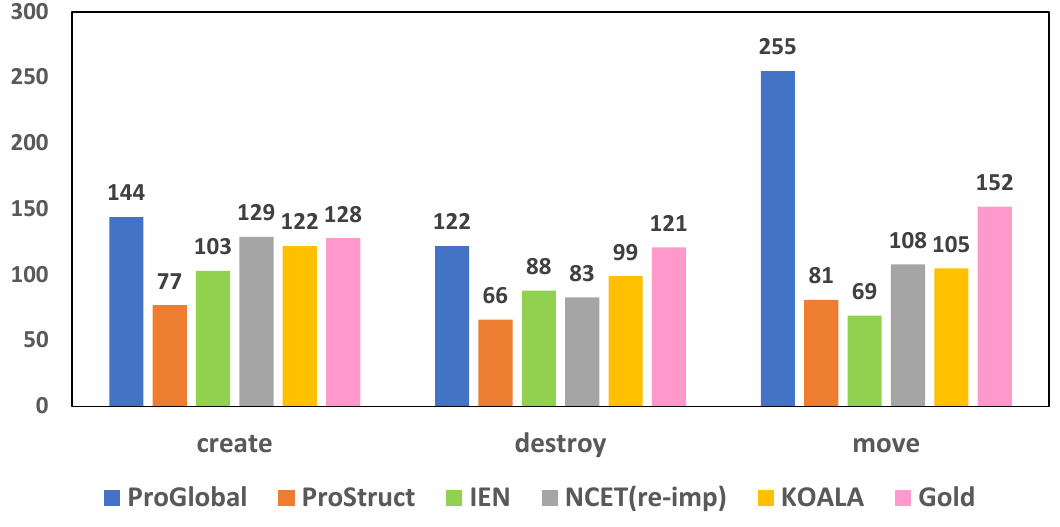}
    \caption{The number of predictions made by models on different state changes. ``Gold'' denotes the ground truth labels.}
    \label{fig:state_type}
\end{figure}

In auxiliary experiments on Recipes, since we re-split the dataset using human-labeled data, we compare \textsc{KoaLa} with its variants and our re-implemented NCET. As shown in Table \ref{tab: recipes}, although not devised for cooking domain (\textit{e.g.}, retrieving ConceptNet triples using recipe ingredients may be noisy), our model still outperforms NCET (9.9\%/13.4\%/10.2\% relative improvements in precision/recall/F1) and other variants in predicting location changes of recipes ingredients, which further proves the effectiveness of our model.

\subsection{Ablations and Analyses}

\subsubsection{Ablation Tests}
In order to further testify the effectiveness of the proposed components in this paper, we perform an ablation test on multiple variants of \textsc{KoaLa}. We remove certain components of our model and test whether it will deteriorate the model's performance. As shown in Table \ref{tab: ablation}, ConceptNet knowledge is proved to be effective even when we simply average their representations (67.7$\rightarrow$68.5). The enhancement brought by attentive knowledge infusion (68.5$\rightarrow$69.2) verifies the efficacy of knowledge selection, which allows the model to select the most relevant knowledge to the input context. Besides, the attention loss contributes to selecting more useful knowledge regarding the prediction target, which leads to another performance upgrade (69.2$\rightarrow$70.4).

\begin{figure*}[tb]
    \centering
    \includegraphics[width=1.0\textwidth]{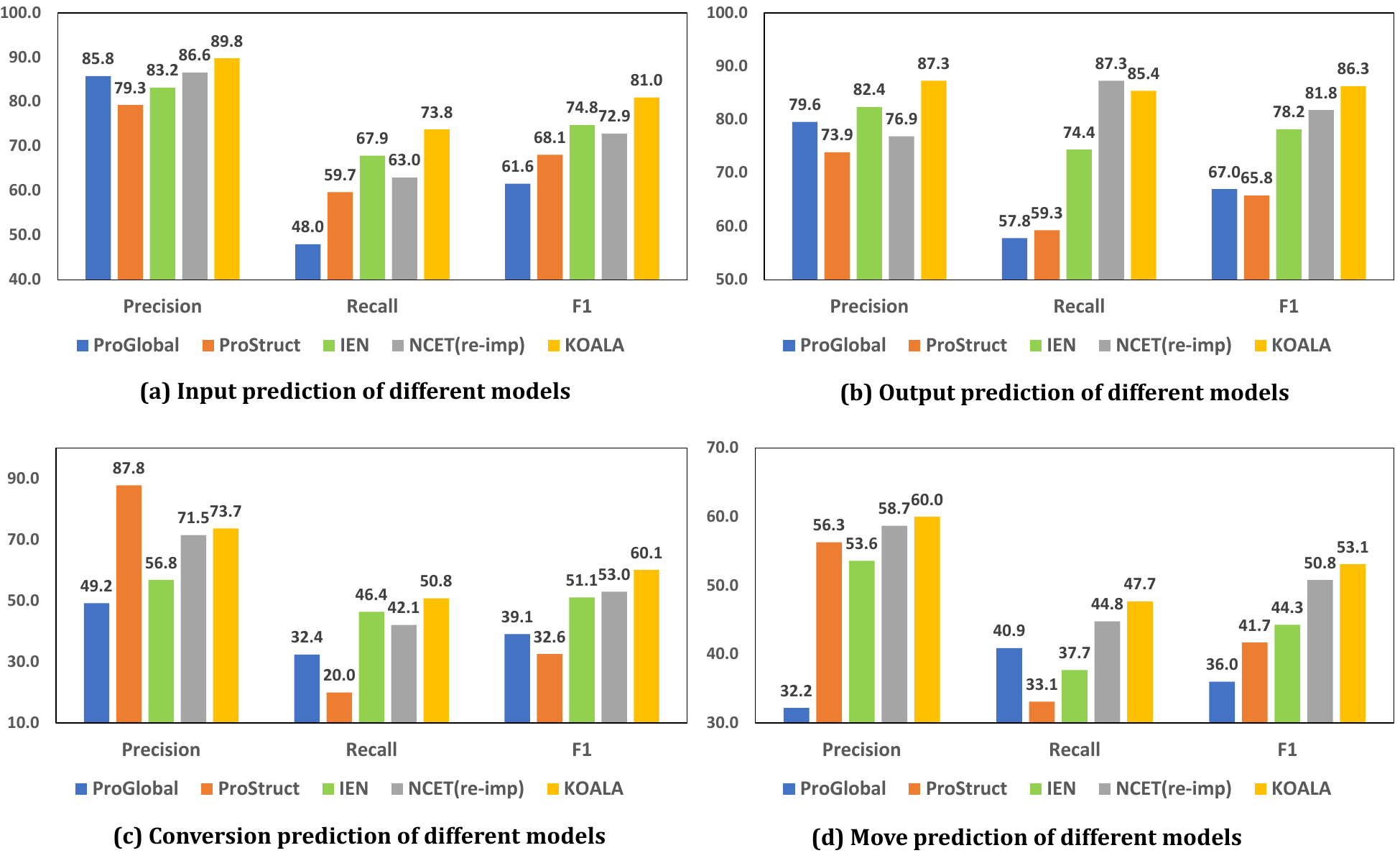}
    \caption{Results of precision/recall/F1 scores of different models on four evaluation aspects of ProPara doc-level task.}
    \label{fig:state_prediction}
\end{figure*}

\begin{figure}[tb]
    \centering
    \includegraphics[width=0.85\linewidth]{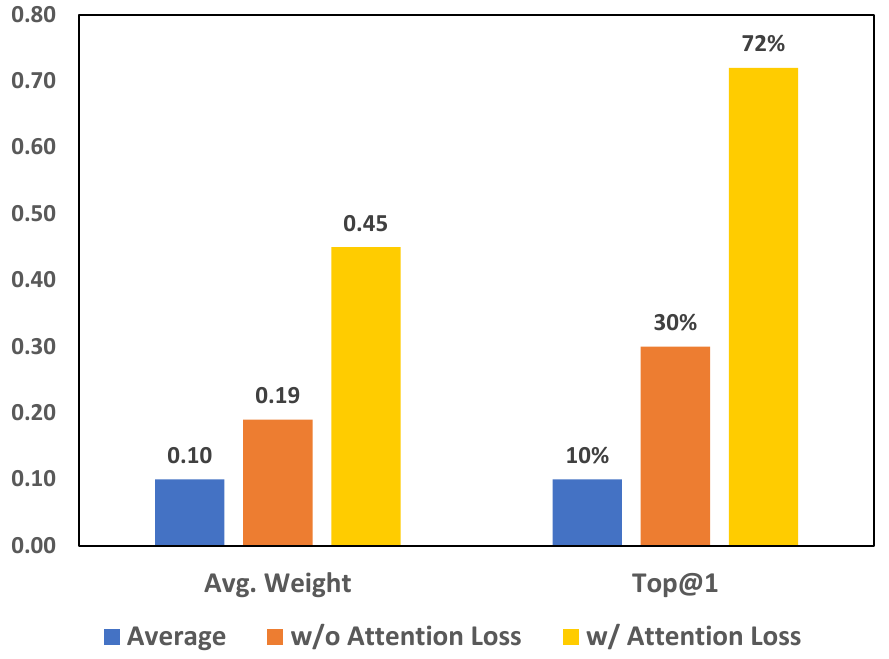}
    \caption{The average attention weights and top-1 percentage of the labeled knowledge triples. Results are collected from the ProPara test set.}
    \label{fig:attn_loss}
\end{figure}

As for multi-stage training, consistent with the ``pre-train then fine-tune'' success in other NLP domains, BERT encoders receive significant performance gain through fine-tuning on the ProPara task (66.5$\rightarrow$68.7). The additional LM fine-tuning phase improves the model for a second time (68.7$\rightarrow$70.4), indicating that pre-fine-tuning on a larger corpus, though in self-supervised manner, strengthens BERT's encoding ability on procedural text. If we remove both ConceptNet knowledge infusion and multi-stage training procedure, the model's performance downgrades to 65.5 F1. Similar results appear in the ablation test on Recipes dataset (Table \ref{tab: recipes}), where we test the effectiveness of ConceptNet knowledge incorporation, BERT fine-tuning as well as the additional LM fine-tuning. Therefore, both ConceptNet knowledge and multi-stage training schema are crucial to \textsc{KoaLa}'s strong performance. ConceptNet triples make the model aware of extra commonsense knowledge to remedy the information insufficiency in some cases, while multi-stage training improves \textsc{KoaLa}'s capability in modeling procedural text.

\subsubsection{Performance in predicting state changes}

To further testify our model's performance in tracking entity states, we decompose the document-level results on ProPara by different types of state changes. Since most previous works did not reveal the detailed evaluation scores of each state type, we compare \textsc{KoaLa} to our re-implemented NCET \cite{gupta2019ncet}, IEN \cite{tang2020ien} and two earlier models ProGlobal \cite{dalvi2018propara} and ProStruct \cite{tandon2018prostruct}.

We first present the detailed results according to four evaluation aspects of the document-level task on ProPara, \textit{i.e.}, inputs, outputs, conversions and moves. As is listed in Figure \ref{fig:state_prediction}, in all aspects, \textsc{KoaLa} shows apparent advantages over baseline systems. Predicting inputs/outputs are easier than the other two targets since models only need to predict the initial and final state of an entity. In both two aspects, \textsc{KoaLa} reaches 80+ F1 scores, suggesting that \textsc{KoaLa}'s ability in answering such coarse-grained questions is approaching maturity. In answering two harder fine-grained questions, conversions and moves, \textsc{KoaLa} also achieves competitive results at around 50\textasciitilde60 F1.

Next, we calculate the total predictions of each state type made by each model. As shown in Figure \ref{fig:state_type}, some models predict either too many or too few state changes. As a result, ProGlobal receives a relatively low precision (61.7) among all previous models due to too many state change prediction; while ProStruct has a relatively low recall (43.0) considering its conservative strategy of predicting fewer state changes. IEN predicts much fewer \texttt{\textbf{create}}s and \texttt{\textbf{move}}s than NCET and \textsc{KoaLa}, resulting in lower recall scores in answering questions about outputs (74.4) and moves (37.7). On the contrary, NCET and \textsc{KoaLa}'s prediction quantity is closer to the ground-truth labels.

Integrating the results from Figure \ref{fig:state_prediction} and \ref{fig:state_type}, we can conclude that: compared to approaches like ProGlobal, ProStruct and IEN, \textsc{KoaLa} predicts neither too many nor too few state changes, leading to relatively high scores in both precision and recall; when comparing to the strong baseline NCET, \textsc{KoaLa} maintains more accurate predictions with a similar quantity of state change predictions.

\begin{figure*}[tb]
    \centering
    \includegraphics[width=1.0\textwidth]{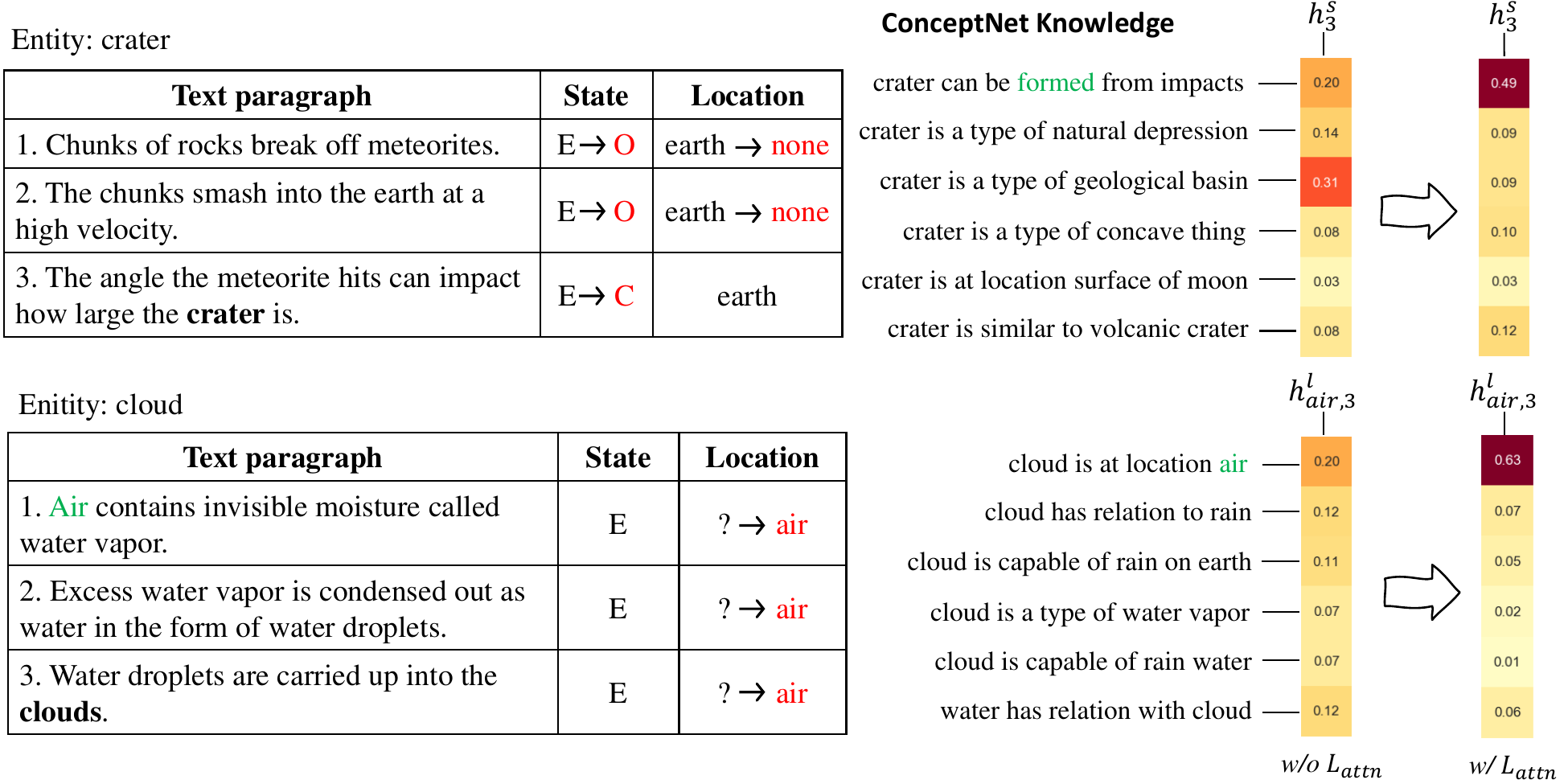}
    \caption{Examples of model predictions \textit{w/} ({\color{red} red}) and \textit{w/o} (black) ConceptNet knowledge. Attention queries in two cases are $\bm{h}^s_3$ and $\bm{h}^l_{air,3}$ respectively. ConceptNet triples are presented as pseudo-sentences. Attention weights \textit{w/} and \textit{w/o} attention loss are visualized as heatmaps, where darker color indicates larger attention weights. We only show part of the retrieved triples due to space limit.}
    \label{fig:case}
\end{figure*}

\subsubsection{Effects of the Attention Loss}

Here, we present the effects introduced by the task-specific attention loss mentioned in Section \ref{sec:attn_loss} on \textsc{KoaLa}'s knowledge selection. For clarity, we calculate those timesteps where only one knowledge triple is labeled as ``relevant'' to the prediction targets on the ProPara test set. As shown in Figure \ref{fig:attn_loss}, the vanilla attention struggles in paying major attention on the labeled knowledge triples. In comparison, the attention loss assists \textsc{KoaLa} in highlighting the labeled triples in the knowledge selection procedure during inference. The average attention weight of the labeled triples increases by 137\% after training with the attention loss. Moreover, such training makes 72\% of the labeled triples become the most attended knowledge during knowledge selection, which is 42\% higher than the vanilla attentive selection (30\%). Therefore, training with attention loss makes \textsc{KoaLa} pay more attention on those knowledge triples which are more statistically probable to be relevant to the prediction targets. Nevertheless, since the attention loss is derived from heuristic labeling, its actual improvements on test results are also bonded to label precision.

\subsubsection{Effects of Multi-Stage Training}

Besides, we also compare the perplexity of the text encoder as an additional evaluation of multi-stage training. Here we use the nouns \& verbs in the test set of ProPara as the evaluation target, because our modified MLM objective during LM fine-tuning is only applied to nouns \& verbs in procedural text corpus. As shown in Table \ref{tab: perplexity}, since ProPara contains many scientific terms which are usually low-frequency nouns in BERT's vocabulary, vanilla BERT has a relatively high perplexity. However, LM fine-tuning on Wikipedia paragraphs largely reduces the perplexity of predicting such tokens. Training BERT on this extended corpus for 1 epoch lowers the perplexity by 51\%, while 5 epochs of training further reduces the perplexity by 64\%. This indicates the fine-tuned BERT encoder performs better in predicting nouns \& verbs, which leads to better token representations. This also shows that the retrieved Wiki paragraphs successfully simulate the writing style of procedural text and covers the terminology of scientific processes. Considering results in Table \ref{tab: recipes}-\ref{tab: perplexity}, training with a larger corpus of procedural text indeed upgrades the model's performance.

\subsection{Case Study}

In Figure \ref{fig:case}, we present two examples in ProPara test set where ConceptNet knowledge assists \textsc{KoaLa} in making correct predictions. We list the predictions made with \& without ConceptNet on the left, and visualize the attention weights assigned to ConceptNet triples while training with \& without attention loss on the right. 

\begin{table}[t]
\resizebox{1.0\columnwidth}!{
\begin{tabular}{@{}l|cccc@{}}
\toprule
\textbf{Epochs}     & pre-trained & 1 epoch & 3 epochs & 5 epochs \\ \midrule
\textbf{Perplexity} & 11.50       & 5.56    & 4.77     & 4.17     \\ \bottomrule
\end{tabular}
}
\caption{Perplexity of the text encoder on nouns \& verbs in ProPara test set during LM fine-tuning. Lower perplexity scores indicate better performances.}
\label{tab: perplexity}
\end{table}

The first case shows how ConceptNet knowledge helps with more accurate state tracking. Although the paragraph does not explicitly state that the crater is created in sentence 3, ConceptNet knowledge tells the model that \textit{``crater can be formed from impacts''}, where ``form'' is a typical verb sign for action \texttt{\textbf{create}}. In fact, ``form'' is included in the co-appearance verb set $\mathcal{V}_{create}$ that we collect from the training data. Although the vanilla attention finds some clues in knowledge triples, it also marks out irrelevant knowledge \textit{``crater is a type of geological basin''}, because $\mathcal{V}_{create}$ has not been applied in training. After given the prompt of co-appearing verbs and trained with the attention loss, the model finally succeeds in paying major attention on the relevant knowledge triple, with its attention weight increasing from 0.20 to 0.49. Benefiting from the corrected state prediction, the model is also able to predict the right location for ``crater'' in step 1 \& 2, since an entity's location before its creation is always ``none''.

In the second case, ConceptNet knowledge mainly helps predict the correct location for entity ``cloud''. In the input paragraph, the entity ``cloud'' and its location ``air'' do not appear in the same step, and their relationship is not mentioned either. Therefore, the model needs extra commonsense knowledge that clouds usually exist in the air. Fortunately, our model locates the relevant knowledge \textit{``cloud is at location air''}, which is extracted from the ConceptNet knowledge graph. Training with attention loss again emphasizes the importance of this knowledge piece, with its attention weight increasing from 0.20 to 0.63. With the help of ConceptNet knowledge and the attention loss, our model is capable of collecting more information from both training data and external knowledge base, leading to more accurate predictions and better performance.

\section{Conclusion and Future Work}
In this work, we propose \textsc{KoaLa}, a novel model for the task of procedural text understanding. \textsc{KoaLa} solves two major challenges in this task, namely commonsense reasoning and data enrichment, by introducing effective methods to leverage external knowledge sources.
Extensive experiments on ProPara and Recipes datasets demonstrate the advantages of \textsc{KoaLa} over various baselines. Further analyses prove that both ConceptNet knowledge injection and multi-stage training contribute to the strong performance of our model. Given the positive results achieved by \textsc{KoaLa}, future work may focus on other issues on procedural text understanding, such as entity resolution or the implicit connection between verbs and states.


\bibliographystyle{ACM-Reference-Format}
\bibliography{ref}

\end{document}